%% file: bare_jrnl.tex
\documentclass[journal]{IEEEtran}

\usepackage{xcolor}
\usepackage{amsmath,amsfonts,amssymb}
\usepackage{graphicx}
\usepackage{booktabs}
\usepackage{multirow}
\usepackage{enumitem}
\usepackage{comment}
\usepackage{arydshln} % for dashed lines in tables
\usepackage{xfrac}
\usepackage{hyperref}

\usepackage{listings}

\definecolor{codegreen}{rgb}{0,0.6,0}
\definecolor{codegray}{rgb}{0.5,0.5,0.5}

\begin{document}
%\title{Designing SAR-ATR Systems for an Open-World Operating Environment} % OLD
\title{A Global Model Approach to Robust Few-Shot SAR Automatic Target Recognition}

% \author{Michael~Shell,~\IEEEmembership{Member,~IEEE,}
%         John~Doe,~\IEEEmembership{Fellow,~OSA,}
%         and~Jane~Doe,~\IEEEmembership{Life~Fellow,~IEEE}% <-this % stops a space
% \thanks{M. Shell was with the Department
% of Electrical and Computer Engineering, Georgia Institute of Technology, Atlanta,
% GA, 30332 USA e-mail: (see http://www.michaelshell.org/contact.html).}% <-this % stops a space
% \thanks{J. Doe and J. Doe are with Anonymous University.}% <-this % stops a space
% \thanks{Manuscript received April 19, 2005; revised August 26, 2015.}}

%\author{Nathan Inkawhich, Matthew Inkawhich, Eric Davis, Uttam %Majumder, Chris Capraro, Yiran Chen, Eric Branch, and John %Nehrbass}%

\author{Nathan Inkawhich\\Air Force Research Laboratory}%
\IEEEaftertitletext{\vspace{-1\baselineskip}} % NAI: prevent unnecessary space

% \author{Jingyang Zhang\textsuperscript{\textdagger}, Nathan Inkawhich\textsuperscript{*}, Randolph Linderman\textsuperscript{\textdagger}, Yiran Chen\textsuperscript{\textdagger}, Hai Li\textsuperscript{\textdagger}\\
% \textsuperscript{\textdagger}Duke University, 
% \textsuperscript{*}Air Force Research Laboratory\\
% {\tt\small jingyang.zhang@duke.edu}}

\markboth{Journal of \LaTeX\ Class Files,~Vol.~14, No.~8, July~2022 (PREPRINT - DRAFT)}%
{Shell \MakeLowercase{\textit{et al.}}: Bare Demo of IEEEtran.cls for IEEE Journals}



\maketitle

\begin{abstract}
In real-world scenarios, it may not always be possible to collect hundreds of labeled samples per class for training deep learning-based SAR Automatic Target Recognition (ATR) models. 
This work specifically tackles the few-shot SAR ATR problem, where only a handful of labeled samples may be available to support the task of interest.
Our approach is composed of two stages. 
In the first, a global representation model is trained via self-supervised learning on a large pool of diverse and unlabeled SAR data. 
In the second stage, the global model is used as a fixed feature extractor and a classifier is trained to partition the feature space given the few-shot support samples, while simultaneously being calibrated to detect anomalous inputs.
Unlike competing approaches which require a pristine labeled dataset for pretraining via meta-learning, our approach learns highly transferable features from unlabeled data that have little-to-no relation to the downstream task.
We evaluate our method in standard and extended MSTAR operating conditions and find it to achieve high accuracy and robust out-of-distribution detection in many different few-shot settings.
Our results are particularly significant because they show the merit of a global model approach to SAR ATR, which makes minimal assumptions, and provides many axes for extendability.
\end{abstract}


% \fi
\IEEEpeerreviewmaketitle

%%%%%%%%%%%%%%%%%%%%%%%%%%%%%%%%%%%%%%%%%%%%%%%%%%%%%%%%%%%
%\newcommand{\G}[1]{{\color{green}{#1}}}
\newcommand{\G}[1]{{\color{codegreen}{#1}}}
\newcommand{\R}[1]{{\color{red}{#1}}}
\newcommand{\M}[1]{\textbf{\color{magenta}{#1}}}
\newcommand{\B}[1]{\textbf{\color{blue}{#1}}}
\newcommand*\rot{\rotatebox{65}}

\input{sections/S1_introduction}

\input{sections/S2_methodology}

\input{sections/S3_experiments}

\input{sections/S4_conclusion}

%\textbf{Disclaimer:} The views expressed in this article are those of the authors and do not reflect official policy of the United States Air Force, Department of Defense or the U.S. Government. Public release number: AFRL-2020-0519

%%%%%%%%%%%%%%%%%%%%%%%%%%%%%%%%%%%%%%%%%%%%%%%%%%%%%%%%%%%

% or


%\section*{Acknowledgment}

\ifCLASSOPTIONcaptionsoff
  \newpage
\fi

\input{sections/appendix}
%%%%%%%%%%%%%%%%%%%%%%%%%%%%%%%%%%%%%%%%%%%% NAI

\end{document}

%% file: sections/S1_introduction.tex
\section{Introduction} \label{sec:intro}

%%% Background

Most modern Synthetic Aperture Radar (SAR) Automatic Target Recognition (ATR) algorithms use Deep Neural Networks (DNNs) to perform the underlying recognition function \cite{tom_book}. 
While DNNs have shown to be very effective in ideal laboratory-like conditions, they present several challenges when being considered for deployment in ``real-world'' scenarios.
One challenge is that DNNs are data-hungry, and often utilize hundreds to thousands of training samples per class to achieve state-of-the-art accuracy.
Another key challenge is that DNNs are notoriously susceptible to producing erroneous predictions on anomalous/out-of-distribution (OOD) inputs (i.e., inputs that fall outside of the categories in the training distribution). 
These challenges spark two main concerns. 
First, what if we, as the model practitioners, cannot obtain hundreds of labeled samples per class for training, and instead can only come up with $\sim$5 (i.e., a few shots)?
And second, in these few-shot settings, does OOD detection become harder because the models are primarily seeking generalization and have not been given enough information to learn highly detailed/nuanced representations of the in-distribution (ID) classes?

%%% Overview Figure
\begin{figure}[t]
  \centering
  %\vspace{-2mm}
  \includegraphics[width=1.0\linewidth]{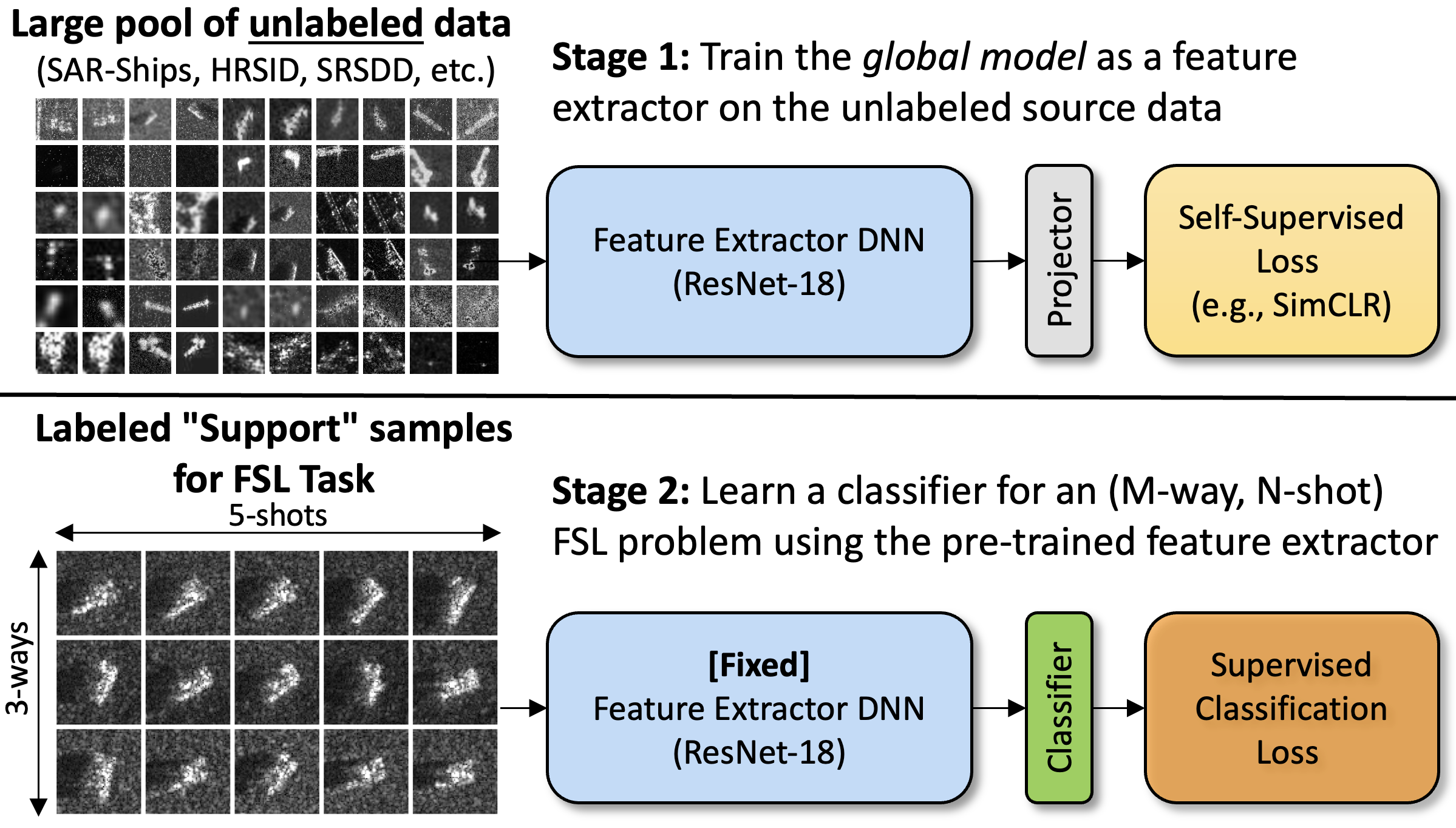}
  \vspace{-5mm}
  \caption{Overview of our two-stage approach. In Stage 1 we train the feature extractor on a pool of unlabeled data and in Stage 2 we learn a classifier on top of the feature extractor for a given few shot learning problem.}
  \vspace{-4mm}
  \label{fig:overview}
\end{figure}

%%% APPROACH / METHOD

Our goal in this paper is to address the stated concerns by developing an accurate and robust SAR target classifier given only a few labeled samples per class. 
We also strive to keep a minimal assumption set to prioritize applicability and extendability.
Our high level approach involves two stages and is shown in Fig.~\ref{fig:overview}. 
In the first stage, a global representation model is trained on a diverse and unlabeled pool of SAR data such that it is capable of extracting quality features from (nearly) any SAR chip. 
In Stage 2, for an arbitrary downstream (M-way, N-shot) few-shot learning (FSL) task we train a light-weight classifier using the global model as a fixed feature extractor.

A key part of our methodology is how we train the global model.
Importantly, we do \textit{not} assume to have labeled data in this step, which eliminates the option for a meta-learning approach \cite{protonet,dktsn}.
To learn quality representations we instead aggregate publicly available data from recent papers and leverage a modern unsupervised learning paradigm called Self-Supervised Learning (SSL). 
Specifically, we use the SimCLR \cite{ChenK0H20} method to represent the SSL class of algorithms in this work.
%SSL algorithms such as SimCLR 
SSL has shown great promise for transfer and few-shot learning in the natural imagery domain \cite{goodembedding_fsl}, but has not been studied widely in the context of SAR ATR.
Our hypothesis is that these modern algorithms can learn a highly flexible~/~expressive~/~useful~/~transferable feature extractor from unlabeled SAR data sourced from different sensors, imaging modes, polarizations, resolutions, and target types. Further, the unlabeled data does not necessarily have to represent the anticipated downstream conditions.
This is contrary to several lines of concurrent research whose goal is to generate high-fidelity synthetic imagery as close to the downstream task conditions as possible for supervised pre-training \cite{sarsim_dataset}.

The second key part of our methodology is how we improve robustness to OOD inputs.
This is accomplished in the classifier training stage and given our stated operating environment it is nearly a ``free-lunch.''
Practically, we can use the SSL model's pretraining dataset as an OOD dataset for outlier exposure (OE) training \cite{HendrycksOE, inkawhich_advoe}. 
Intuitively, the classifier is taught to be accurate and confident on labeled task data while being minimally confident on the OOD data.
This objective creates a confidence calibration effect which can be exploited to detect novel OOD samples during deployment \cite{inkawhich_onlineOOD}.

%%% Experiments & Contributions

Using MSTAR \cite{Ross1998MSTAR} as our primary few-shot test environment, we find that our methodology can generate highly accuracy models in both standard and extended operating conditions.
We also find that by OE training in Stage 2, our models can reliably detect a spectrum of fine- to coarse-grained OOD types unseen during training. 
Finally, we uncover and discuss an important trade-off between OOD generalization and detection. 
Overall, we make the following contributions:
\begin{itemize}
    \item We show a proof-of-concept that a highly transferable global SAR feature extractor can be trained without labels on diverse data and used to great effect in downstream FSL tasks;
    \item We develop a robust FSL classification scheme that can reliably detect and reject OOD inputs at test time without adding overhead or assumptions;
    \item We uncover a critical trade-off between OOD detection and generalization that directly motivates future work.
\end{itemize}

%%% Random thoughts:

%- Few-shot SAR-ATR is also particularly challenging because the target signatures can change drastically given minimal shifts in viewing angle and any target motion.

%- Mention that we are not aware of many other FSL approaches that do not require labeled pretraining data from a domain relevant dataset.

%%% End

%% file: sections/S2_methodology.tex
\section{Methodology}

%%%%%%%%%%%%%%%%%%%%%%%%%%%%%%%%%%%%%%%%%%%%%%%%%%%%%%%%%%%%%
\subsection{Global Model Training}

%%%%%%%%%%%%%%%%%%%%%%%%%%%%%%%
\subsubsection{Data Collection}

\begin{figure}[t]
  \centering
  %\vspace{-2mm}
  \includegraphics[width=0.96\linewidth]{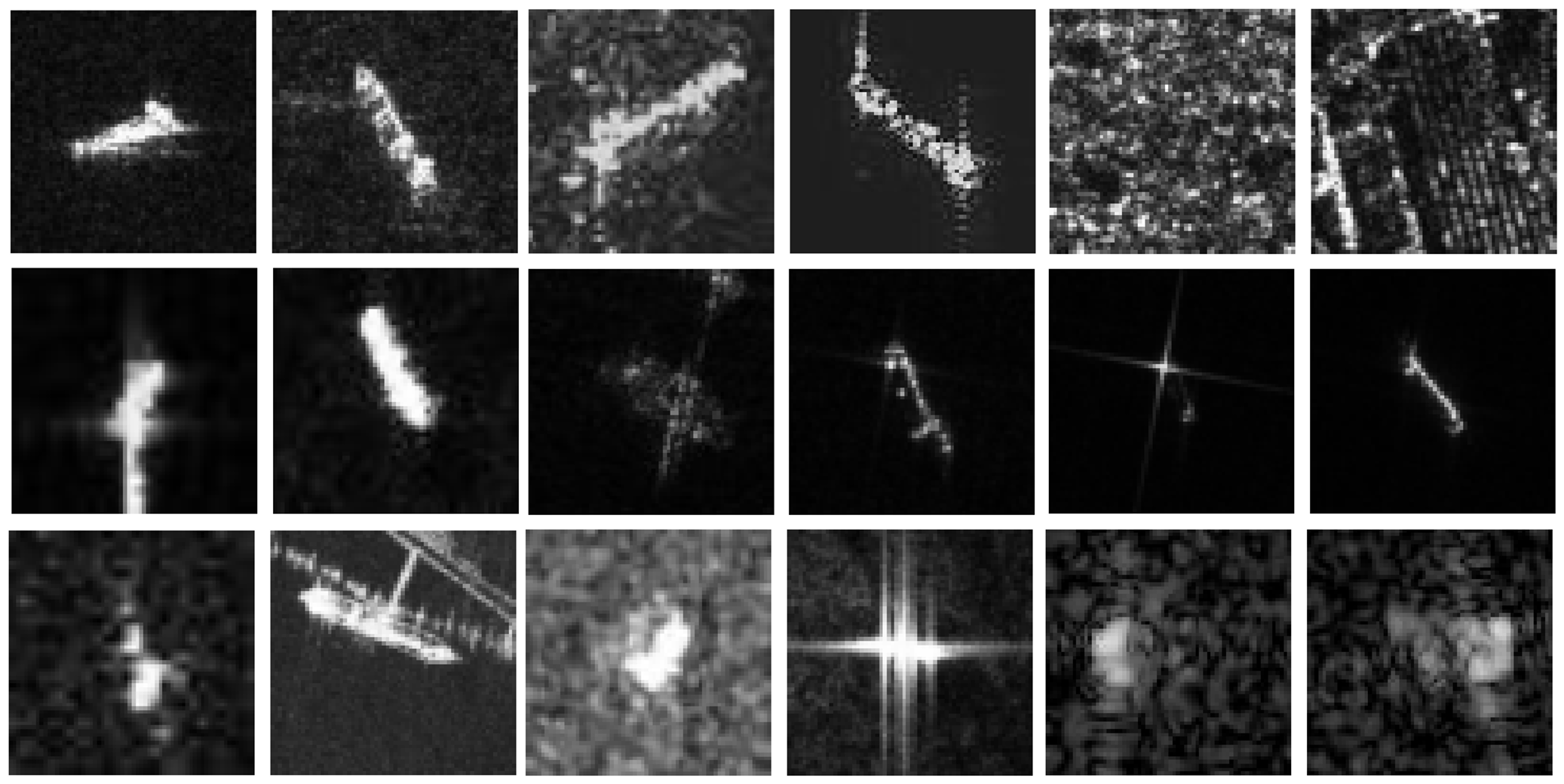}
  \vspace{-1mm}
  \caption{Samples from $\mathcal{D}_{pretrain}$, sourced from: SAR-Ships \cite{sarship_dataset}; HRSID \cite{hrsid_dataset}; FUSAR \cite{fusar_dataset}; SSDD \cite{ssdd_dataset}; LS-SSDD \cite{lsssdd_dataset}; Dual-Pol Ships \cite{dualpolships_dataset}; SRSDD \cite{srsdd_dataset}; and CVDome \cite{cvdome_dataset}.}
  \vspace{-1mm}
  \label{fig:pretrain_data}
\end{figure}

Our global model approach is highly data driven. 
In order to learn a generalized and transferable SAR feature extractor, intuition says we must train it on data from different sensors, imaging modes, polarizations, resolutions, target/scene types, etc.
Thus, the first step in our methodology is to aggregate data from the following public sources: SAR-Ships \cite{sarship_dataset}; HRSID \cite{hrsid_dataset}; FUSAR \cite{fusar_dataset}; SSDD \cite{ssdd_dataset}; LS-SSDD \cite{lsssdd_dataset}; Dual-Pol Ships \cite{dualpolships_dataset}; SRSDD \cite{srsdd_dataset}; and CVDome \cite{cvdome_dataset}.
After the necessary pre-processing (e.g., chipping from a full frame), we are left with $\sim$100k unlabeled SAR chips which we refer to as $\mathcal{D}_{pretrain}$.
For reference, Fig.~\ref{fig:pretrain_data} shows a handful of samples from this set. 
An important note is that $90\%$ of $\mathcal{D}_{pretrain}$ chips contain ship-like targets and \textit{none} of the chips contain MSTAR land-vehicle targets. 
Given that MSTAR will be our downstream task for FSL evaluation, this sets up for a true test of generalization.
However, in practice if data resembling the anticipated downstream task were available it could easily be leveraged here.

%%%%%%%%%%%%%%%%%%%%%%%%%%%%%%%
\subsubsection{Representation Learning}

With $\mathcal{D}_{pretrain}$ curated we can execute Stage 1, which is to train the global model feature extractor $f_\phi$. 
We use the powerful SimCLR \cite{ChenK0H20} contrastive learning algorithm, which trains by enforcing that two ``views'' of the same image lie near each-other in a feature space, while views of different images lie far away.
Each view is created via aggressive stochastic augmentations that preserve the underlying semantic features but force the model to learn an informative feature set via minimizing the contrastive objective.  
We specifically leverage the normalized temperature-scaled cross-entropy loss (NT-Xent) from \cite{ChenK0H20}, and refer the reader to that work for a more technical description of the algorithm and \cite{simclr_github} for a PyTorch implementation.

Perhaps the most unique part about the application of SimCLR (and many other SSL algorithms) to the SAR ATR setting is the composition of augmentations needed to create informative views. 
Since the SimCLR algorithm was developed and tested in the natural imagery domain, the standard augmentation pipeline contains transforms that are relevant to RGB images. 
However, several of the key augmentations such as color jittering and random gray-scaling \cite{simclr_github} do not fit with our magnitude-only gray-scale SAR modality.
Thus, we create a new augmentation pipeline made of seven image-level transformations that work to create translation invariance, and to manipulate the signal-to-noise ratio and frequency content of the chips. 
Included are random resized cropping, random horizontal flipping, element-wise power-scaling, Gaussian noising and filtering, and linear re-scaling.
See Supplemental A for more information about our transformation pipeline.
Lastly, we mention that we do not believe our design choices made in this section are ``optimal.'' 
Important areas of future work are to experiment with other SSL algorithms (see Supplemental C) and to develop more domain-relevant augmentations that may lead to higher quality learned features.

%%% D_pretrain Examples Figure

%%% MSTAR SOC Accuracy Table
\input{tables/table_mstar_soc}
\subsection{Classifier Training}

In Stage 2 (ref. Fig.~\ref{fig:overview}) we use the global model $f_\phi$ as a fixed feature extractor and learn a classifier $c_\theta$ for the given (M-way, N-shot) FSL task, described by $\mathcal{D}_{support}=\{(x_i,y_i)\}_{i=1 \ldots (M \times N)}$.
To be clear, ``ways'' refers to the number of categories in the label-space and ``shots'' is number of labeled training chips per category. 
Unlike related works in the natural imagery domain that use simple linear regression or nearest neighbor-style classifiers \cite{goodembedding_fsl}, we find it beneficial to utilize a 2-layer neural network which has the flexibility to learn non-linear decision boundaries.

We specifically consider two ways to train $c_\theta$. 
The first, called the \textit{basic} method, is to train under a vanilla supervised learning objective which minimizes empirical risk on $\mathcal{D}_{support}$ samples only. 
Mathematically, the basic training method is described as
\begin{equation}
\min_{\theta} \mathop{\mathbb{E}}_{(x,y)\sim\mathcal{D}_{support}}\big [ L(c_{\theta}(f_{\phi}(x)), y) \big ], 
\end{equation}
where $L$ is the cross-entropy loss between the classifier's prediction and the ground truth label $y$.
The intuition for this training method is to learn a classifier that produces accurate and confident predictions on the ID data.

While the basic method can achieve high ID accuracy, it gives no guidance to the model for how to behave when an OOD input is encountered. 
Without violating or adding any assumptions, we can do something clever to greatly improve the classifier's ability to handle OOD inputs. 
Specifically, we can re-purpose the $\mathcal{D}_{pretrain}$ dataset from Stage 1 as an \textit{outlier exposure} (OE) set in Stage 2.
The intuition for OE \cite{HendrycksOE} is to teach the model a confidence calibration on ID and OOD inputs -- it should make accurate and confident predictions on ID data and minimally confident predictions on OOD data.
Functionally, the minimum confidence state is when the predicted probability is \sfrac{1}{\# classes}, so to achieve the desired effect we set the training targets of the OE data to be a Uniform distribution over the classes $\mathcal{U}_{\mathcal{C}}$, while the targets of the ID data remain 1-hots.
The complete OE objective is
\begin{multline}
   \min_{\theta} \mathop{\mathbb{E}}_{(x,y)\sim\mathcal{D}_{support}}\big [ L(c_{\theta}(f_{\phi}(x)), y) \big ] \\ + \lambda \mathop{\mathbb{E}}_{\tilde{x}\sim\mathcal{D}_{pretrain}}\big [ L(c_{\theta}(f_{\phi}(\tilde{x})), \mathcal{U}_{\mathcal{C}}) \big ], 
\end{multline}
where $\lambda$ is a weighting factor between the ID- and OOD-focused terms that we set to 0.5 here \cite{HendrycksOE}.

%%%%%%%%%%%%%%%%%%%%%%%%%%%%%%%%%%%%%%%%%%%%%%%%%%%%%%%%%%%%%
\subsection{OOD Detection}

The final detail of our methodology is how we detect OOD samples during deployment.
To conceptualize the process, think about each test input $x$ being assigned a real-valued score $\mathcal{S}_{ID}(x)$ as a measure of its ID-ness.
During operation, if $\mathcal{S}_{ID}(x) \geq \beta_{thresh}$ (an application dependent threshold), then the input would be considered ID and the classifier would release its prediction over the set of known classes. Else, if $\mathcal{S}_{ID}(x) < \beta_{thresh}$, the input is considered OOD and the system abstains from releasing the prediction.
In this work we use a temperature-scaled Maximum Softmax Probability detector \cite{odin_ood} to produce $\mathcal{S}_{ID}(x)$ scores, described as
\begin{equation}
\mathcal{S}_{ID}(x) = \max_{i\in\mathcal{C}} \frac{exp \big( c_{\theta}^{(i)}(f(x))/\tau \big )}{\sum_{j}^{\mathcal{|C|}} exp \big( c_{\theta}^{(j)}(f(x))/\tau \big )}.
\end{equation}
Here, $\tau$ is a temperature hyperparameter which we set to 100 via validation. 
The intuition for this detection scheme is straightforward: ID samples should be predicted by the classifier with higher confidence than OOD samples. 
This fits naturally with the OE training goal, making it a logical choice in our system.
We note that OOD detection is currently a popular research topic and future works may investigate alternative scoring mechanisms.

%% file: tables/table_mstar_soc.tex
%%%%%%%%%%%%%%%%%%%%%%%%%%%%%%%%%%%%%%%%%%%%%%% MSTAR SOC RESULTS -- NO ROTATED LABELS
\begin{table*}[t]
\centering
\caption{MSTAR Standard Operating Condition Accuracy Results (with 95\% confidence interval)}
\vspace{-1mm}
\label{tab:mstar_soc}
\resizebox{1.\textwidth}{!}{
\begin{tabular}{lccccccccccccccc}
\toprule
                 & \multicolumn{5}{c}{2-way}                     & \multicolumn{5}{c}{5-way}                     & \multicolumn{5}{c}{10-way}                    \\ \cmidrule(lr){2-6} \cmidrule(lr){7-11} \cmidrule(lr){12-16}
Method           & {1-shot} & {5-shot} & {10-shot} & {20-shot} & {25-shot} & {1-shot} & {5-shot} & {10-shot} & {20-shot} & {25-shot} & {1-shot} & {5-shot} & {10-shot} & {20-shot} & {25-shot} \\ \midrule
Scratch-AConv    &  59.6\scriptsize{$\pm$1.2}  & 73.6\scriptsize{$\pm$1.2}   & 84.3\scriptsize{$\pm$0.9}    & 92.1\scriptsize{$\pm$0.7}    & 93.8\scriptsize{$\pm$0.6}    & 31.6\scriptsize{$\pm$0.6}   & 55.9\scriptsize{$\pm$0.8}   & 72.3\scriptsize{$\pm$0.8}    & 85.7\scriptsize{$\pm$0.5}    & 88.5\scriptsize{$\pm$0.4}    & 20.9\scriptsize{$\pm$0.3}   & 47.4\scriptsize{$\pm$0.4}   & 65.0\scriptsize{$\pm$0.4}    & 80.7\scriptsize{$\pm$0.2}    & 84.5\scriptsize{$\pm$0.2}    \\
Scratch-RN18     &  55.8\scriptsize{$\pm$1.2}  & 74.9\scriptsize{$\pm$1.4}   & 84.1\scriptsize{$\pm$1.1}    & 89.6\scriptsize{$\pm$0.8}    & 92.1\scriptsize{$\pm$0.6}    & 30.6\scriptsize{$\pm$0.7}   & 53.7\scriptsize{$\pm$0.8}   & 67.0\scriptsize{$\pm$0.7}    & 79.3\scriptsize{$\pm$0.6}    & 83.2\scriptsize{$\pm$0.5}    & 20.7\scriptsize{$\pm$0.3}   & 43.4\scriptsize{$\pm$0.4}   & 58.9\scriptsize{$\pm$0.4}    & 74.0\scriptsize{$\pm$0.4}    & 78.5\scriptsize{$\pm$0.3}    \\ [0.2mm] \cdashline{1-16} \\[-2.3mm]
SimCLR+basic     &  64.5\scriptsize{$\pm$1.4}  & 83.0\scriptsize{$\pm$1.1}   & 91.4\scriptsize{$\pm$0.7}    & 96.1\scriptsize{$\pm$0.4}    & 97.4\scriptsize{$\pm$0.2}    & 38.8\scriptsize{$\pm$0.7}   & 68.4\scriptsize{$\pm$0.7}   & 82.6\scriptsize{$\pm$0.5}    & 91.8\scriptsize{$\pm$0.3}    & \textbf{94.0}\scriptsize{$\pm$0.2}    & 28.3\scriptsize{$\pm$0.3}   & 59.2\scriptsize{$\pm$0.3}   & 75.2\scriptsize{$\pm$0.2}    & \textbf{87.7}\scriptsize{$\pm$0.1}    & \textbf{90.6}\scriptsize{$\pm$0.1}    \\
SimCLR+OE        &  \textbf{65.4}\scriptsize{$\pm$1.5}  & \textbf{86.4}\scriptsize{$\pm$1.0}   & \textbf{92.2}\scriptsize{$\pm$0.8}    & \textbf{96.5}\scriptsize{$\pm$0.4}    & \textbf{97.5}\scriptsize{$\pm$0.2}    & \textbf{39.9}\scriptsize{$\pm$0.7}   & \textbf{69.7}\scriptsize{$\pm$0.6}   & \textbf{83.3}\scriptsize{$\pm$0.5}    & \textbf{92.3}\scriptsize{$\pm$0.3}    & 93.8\scriptsize{$\pm$0.2}    & \textbf{28.8}\scriptsize{$\pm$0.3}   & \textbf{60.5}\scriptsize{$\pm$0.3}   & \textbf{75.6}\scriptsize{$\pm$0.2}    & 87.0\scriptsize{$\pm$0.1}    & 89.3\scriptsize{$\pm$0.1}    \\ \bottomrule
\end{tabular}
}
\vspace{-2.5mm}
\end{table*}

%% file: sections/S3_experiments.tex
\section{Experiments}

To show the merit of our approach to the robust few-shot SAR ATR problem we perform three primary experiments and one analysis. 
The first experiment (Sec.~\ref{sec:soc}) shows results in MSTAR's Standard Operating Condition (SOC) and the second (Sec.~\ref{sec:eoc}) investigates generalization performance in an Extended Operating Condition (EOC). 
The third experiment (Sec.~\ref{sec:ood}) measures OOD detection performance against a spectrum of granularities and the analysis (Sec.~\ref{sec:tradeoff}) discusses a critical trade-off between OOD detection and generalization.

All of the experiments follow a similar setup.
In Stage 1, we initialize the feature extractor as a ResNet-18 (RN18) \cite{resnet} backbone with output dimension 512, and the Projector network as a 3-layer neural net \cite{simclr_github} with output dimension 128. 
These components are trained with an ADAM optimizer for 200 epochs, with 1024 batch size, weight decay of 1e-4, and a learning rate of 3e-4 following a cosine decay schedule. 
For the SimCLR NT-Xent loss we use a temperature of 0.01. 
In Stage 2, we discard the Projector network, fix the weights of the RN18, and initialize our 2-layer neural net classifier. 
We train the classifier for 500 iterations, with label smoothing, using an ADAM optimizer with cosine decayed learning rate starting at 1e-3. 
On the data side, we train with $64\times64$~px crops and use Gaussian noise and random flipping augmentations.
Finally, the results in this document are averages over 250 randomized runs.

% SimCLR training: 
% - RN18 architecture
% - bsize=1024
% - lr = 3e-4 with cos decay
% - ADAM optimizer with weight\_decay=1e-4
% - num\_epochs = 200
% - projection dimension = 128
% - NT-Xent temp = 0.01
% Classifier Training
% - 500 iters
% - 5x OE bsize
% - lblsmoothing=0.1
% - centercrop=64
% - ADAM optimizer with lr=1e-3 and cosine decay
% - random hflip
% - gaus noise = 0.1
%All numbers we report are computed averages over 250 randomized runs 

%%%%%%%%%%%%%%%%%%%%%%%%%%%%%%%%%%%%%%%%%%%%%%%%%%%%%%%%%%%%%
\subsection{Standard Operating Conditions} \label{sec:soc}

To measure SOC performance on MSTAR we train the models on 17$^{\circ}$ elevation imagery and test the accuracy of classifying 15$^{\circ}$ data \cite{aconvnet}.
As mentioned, because we do not assume to have labeled pretraining data, meta-learning-based FSL methods are inappropriate for comparison (also see Supplemental B). 
For baselines we instead train supervised models from scratch on the $\mathcal{D}_{support}$ samples only. 
We examine two architectures of varying complexities: A-ConvNet \cite{aconvnet} and RN18 (displayed as Scratch-\{AConv,RN18\}).
The MSTAR SOC results are shown in Table~\ref{tab:mstar_soc} for \#-ways=\{2,5,10\} and \#-shots=\{1,5,10,20,25\}, covering a gamut of potential scenarios.

To start, we find several intuitive takeaways. 
First, accuracy increases with the number of shots; and second, as the number of ways increases the models need more shots to achieve the same performance. 
In all cases, our SimCLR-based models significantly outperform the models trained from scratch, and in most cases the OE classifier has a slight edge over the basic classifier. 
When only given 10-shots, our average margin of improvement over the best baseline is +9.8\%.
On full 10-class MSTAR, we reach 90\% average accuracy at $\sim$25-shots, which is a 6\% improvement over the best baseline.
Lastly, we want to emphasize the implications of these results. 
The self-supervised SimCLR backbone, which has been trained on mostly ships, is able to separate the MSTAR data by class even though it has never been trained on such data!

%%%%%%%%%%%%%%%%%%%%%%%%%%%%%%%%%%%%%%%%%%%%%%%%%%%%%%%%%%%%%
\subsection{Extended Operating Conditions} \label{sec:eoc}

For the MSTAR EOC test we train the classification models on 17$^{\circ}$ elevation imagery and test on 30$^{\circ}$ data \cite{aconvnet}. 
The large difference in collection geometry causes a sizable change in the target signatures and offers a more challenging test of generalization.
Our results are reported in Table~\ref{tab:mstar_eoc}.
We see several similar trends to the SOC results w.r.t. the \#-shots and \#-ways.
In all cases, the SimCLR-based models outperform the train-from-scratch baselines, while the basic and OE classifiers perform very similarly to each other.
Interestingly, the largest margins over the baselines are at the lowest number of shots. For example, at 5-shots our average improvement over the best baseline is +9\%.

%%% MSTAR EOC Accuracy Table
\input{tables/table_mstar_eoc}

%%%%%%%%%%%%%%%%%%%%%%%%%%%%%%%%%%%%%%%%%%%%%%%%%%%%%%%%%%%%%
\subsection{Out-of-Distribution Detection} \label{sec:ood}

\begin{figure}[t]
  \centering
  %\vspace{-2mm}
  \includegraphics[width=1.0\linewidth]{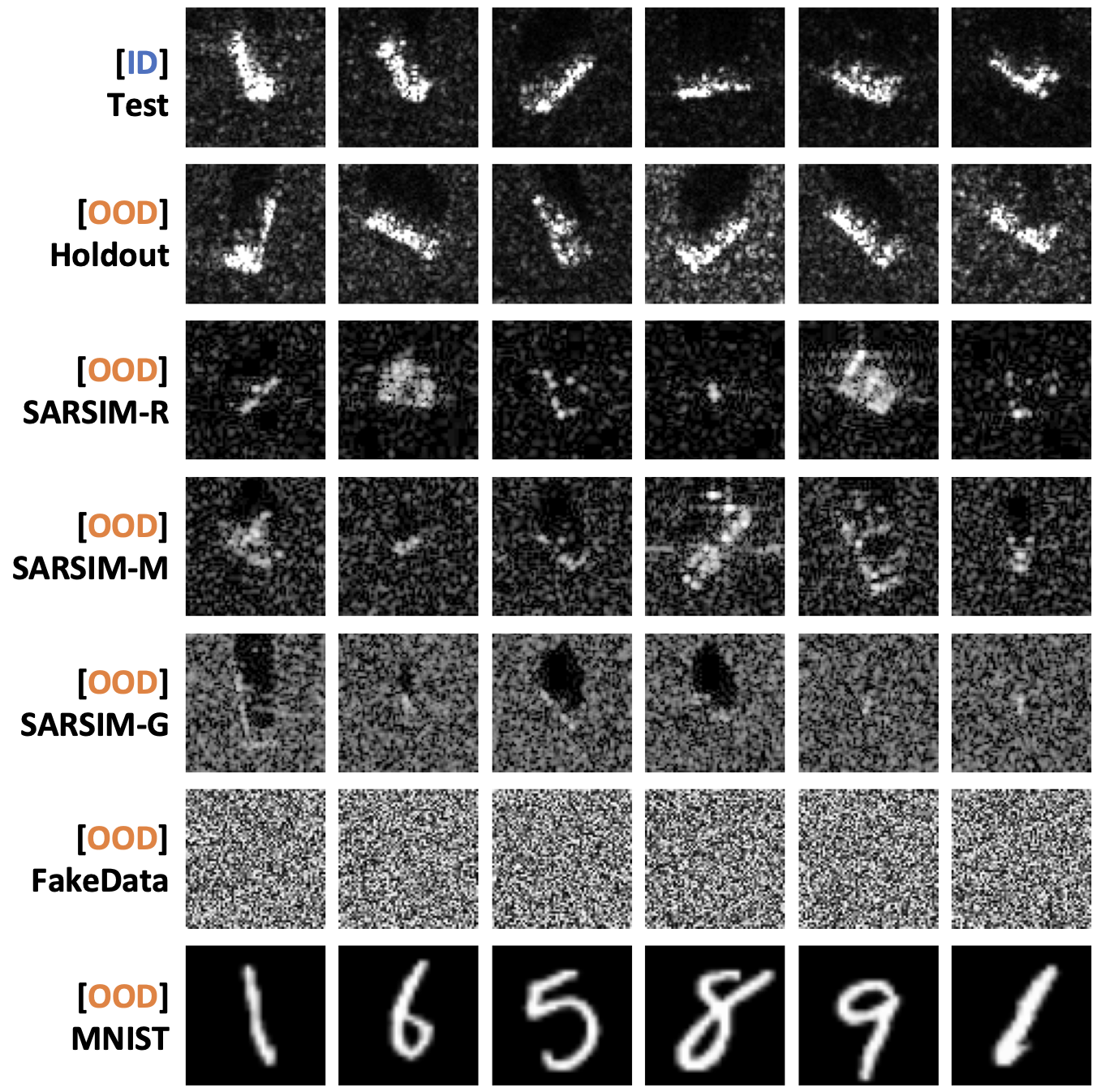}
  %\vspace{-3mm}
  \caption{Sample test chips from each dataset used in the OOD experiment.}
  \label{fig:ood_data}
\end{figure}

Our goal in the third experiment is to examine OOD detection performance across a spectrum of difficulties/granularities.
To do this we leverage OOD data from various sources. 
For a particularly hard OOD test we use a holdout scheme. 
Starting with a 7-way classifier, where in each test iteration 7 of the 10 MSTAR classes are randomly selected for use in the ID label space, we use data from the remaining 3 classes as \textit{Holdout} OOD samples. 
For medium difficulty OOD, we use data from \textit{SARSIM-Roads}, \textit{SARSIM-Medium}, and \textit{SARSIM-Grass} \cite{sarsim_dataset}. 
This is synthetic SAR data which has been generated to specifically match the collection conditions of MSTAR with different backgrounds (note, we remove the Bulldozer and Tank classes to avoid potential ID/OOD ambiguities).
For coarse-grained OOD we generate \textit{FakeData} by creating random Uniform noise chips and use the \textit{MNIST} test set which contains hand-written digits. Both are clearly OOD w.r.t. any SAR classifier.
Fig.~\ref{fig:ood_data} shows examples from each of these datasets. Importantly, keep in mind that none of the test OOD data is in $\mathcal{D}_{pretrain}$ (the OE set), meaning the model has never seen it before.

Table~\ref{tab:mstar_ood} shows the average OOD detection performance (as measured by \% AUROC \cite{odin_ood}) of a 7-way MSTAR SOC classifier at different \#-shots.
Note, the AUROC value can be interpreted as the probability that an ID test input would have a greater $\mathcal{S}_{ID}$ score than an OOD input \cite{smaxthresh}, where 100\% is a perfect detector.
At each setting we examine the performance of the basic versus OE classifier, which we know both achieve high ID accuracy from Sec.~\ref{sec:soc}.
The most important takeaway from Table~\ref{tab:mstar_ood} is that the OE classifier is better than the basic classifier in \textit{all} cases by a large margin.
Interestingly, for the basic model the coarse-grained OOD sets are the hardest to detect, while the OE classifier detects these almost perfectly.
This surprising behavior is something we plan to study in a future work.
Another interesting finding is that detection of \textit{Holdout} samples improves significantly as \#-shots increases, while detection of the other OOD types stays relatively constant w.r.t.~\#-shots. 
Finally, a somewhat expected result is that for OE models, the \textit{Holdout} samples are the most difficult to detect, while the medium- and coarse-grained OOD inputs are less challenging.
Given that these results still show room for improvement, we highly encourage future system designers (whether in few-shot scenarios or not) to use OOD-aware classifier training schemes such as OE to build robustness to the many forms of OOD inputs.

%%% OOD Examples Figure

%%% MSTAR OOD Table
\input{tables/table_mstar_ood}

%%%%%%%%%%%%%%%%%%%%%%%%%%%%%%%%%%%%%%%%%%%%%%%%%%%%%%%%%%%%%
\subsection{Generalization vs. Detection Trade-off} \label{sec:tradeoff}

\begin{figure}[t]
  \centering
  %\vspace{-2mm}
  \includegraphics[width=1.0\linewidth]{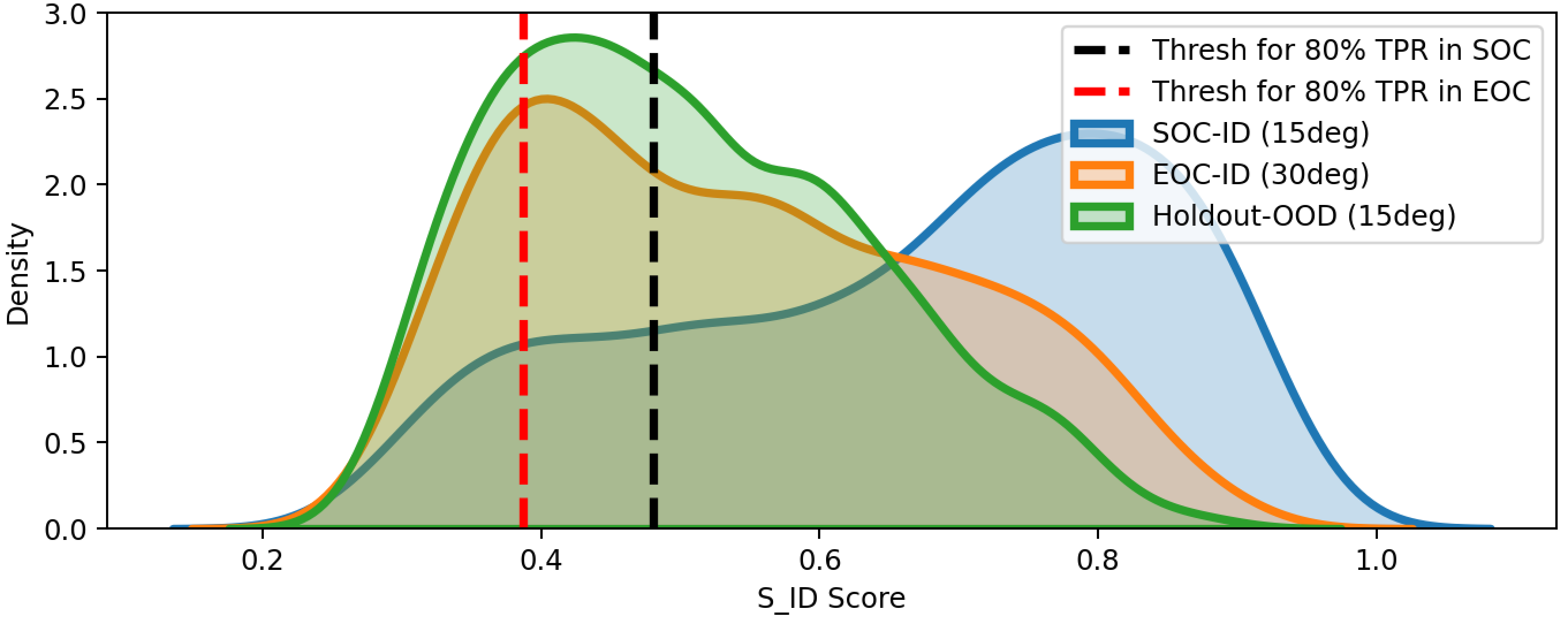}
  %\vspace{-3mm}
  \caption{Density plots of $\mathcal{S}_{ID}$ scores for ID data in SOC, ID data in EOC, and Holdout OOD data. Also shown are hypothetical thresholds to achieve 80\% True Positive Rates on ID data in SOC and EOC.}
  \label{fig:density_plot}
\end{figure}

Our final experiment is an analysis that deliberates the question: can we have it all? 
Specifically, can we have high accuracy on ID data in SOC \textit{and} high accuracy on ID classes in EOC \textit{and} reliable OOD detection?
To answer this question we create Fig.~\ref{fig:density_plot}, which shows density plots of $\mathcal{S}_{ID}(x)$ scores measured on a SimCLR+OE model for ID test data in SOC ({\color{blue}{SOC-ID}}), ID test data in EOC ({\color{orange}{EOC-ID}}), and Holdout OOD data in SOC ({\color{codegreen}{Holdout-OOD}}).

Firstly, we observe that the {\color{blue}{SOC-ID}} and {\color{codegreen}{Holdout-OOD}} are relatively separable, which matches the findings in Table~\ref{tab:mstar_ood}.
However, the {\color{orange}{EOC-ID}} and {\color{codegreen}{Holdout-OOD}} are not very separable, as both are predicted with relatively low scores. 
Thus, with our method if you require good {\color{codegreen}{Holdout-OOD}} detection then most of the {\color{orange}{EOC-ID}} would also get flagged as OOD.
Conversely, if you require high recall on {\color{orange}{EOC-ID}} then many {\color{codegreen}{Holdout-OOD}} would pass through the detector.
These results indicate that we \textit{cannot} have it all with this design, and highlight a serious trade-off to be considered in future work.
Finally, while this trade-off may not be completely unique to the FSL setting, our previous observation that less shots makes Holdout-OOD detection harder (Sec.~\ref{sec:ood}) means this trade-off may be more dire in the few-shot setting.

%%% Generalization vs Detection Density Plot Figure

%% file: tables/table_mstar_eoc.tex
%%%%%%%%%%%%%%%%%%%%%%%%%%%%%%%%%%%%%%%%%%%%%%% MSTAR EOC RESULTS
\begin{table}[t]
\centering
\caption{MSTAR Extended Operating Condition Accuracy Results}
\label{tab:mstar_eoc}
\resizebox{1.\linewidth}{!}{
\begin{tabular}{lcccccccc}
\toprule
                 & \multicolumn{4}{c}{2-way}                     & \multicolumn{4}{c}{4-way}                    \\ \cmidrule(lr){2-5} \cmidrule(lr){6-9}
Method           & \rot{5-shot} & \rot{10-shot} & \rot{20-shot}  & \rot{25-shot} & \rot{5-shot} & \rot{10-shot} & \rot{20-shot} & \rot{25-shot}  \\ \midrule
Scratch-AConv    &  68.9  & 78.4   & 87.0    & 89.3    & 53.8    & 66.2   & 74.3   & 76.3        \\
Scratch-RN18     &  72.1  & 82.1   & 86.6    & 86.9    & 56.0    & 65.8   & 73.2   & 74.5        \\ [0.2mm] \cdashline{1-9} \\[-2.3mm]
SimCLR+basic     &  78.8  & 85.4   & \textbf{90.3}    & \textbf{90.7}    & 65.2    & \textbf{74.2}   & \textbf{80.5}   & \textbf{82.0}        \\
SimCLR+OE        &  \textbf{80.9}  & \textbf{85.7}   & 89.6    & 89.8    & \textbf{65.3}    & 74.1   & 79.7   & 81.0        \\

\bottomrule
\end{tabular}
}
\end{table}

%% file: tables/table_mstar_ood.tex
%%%%%%%%%%%%%%%%%%%%%%%%%%%%%%%%%%%%%%%%%%%%%%% MSTAR OOD DETECTION RESULTS (copy)
\begin{table}[t]
\centering
\caption{MSTAR Out-of-Distribution Detection Results for 7-way classifier (\% AUROC)}
\label{tab:mstar_ood}
\resizebox{\linewidth}{!}{
\begin{tabular}{lccccc}
\toprule
             & \multicolumn{5}{c}{Notation = basic / OE} \\ \cmidrule(lr){2-6}
OOD Set      & 1-shot & 5-shot & 10-shot & 20-shot & 25-shot \\ \midrule
Holdout      & 54.1 / \textbf{56.0}        & 64.1 / \textbf{67.2}        & 70.2 / \textbf{76.2}        & 78.0 / \textbf{81.9}        & 81.7 / \textbf{84.0}        \\
SARSIM-R & 59.6 / \textbf{93.2}        & 56.3 / \textbf{95.9}        & 58.8 / \textbf{96.6}        & 66.0 / \textbf{97.2}        & 68.3 / \textbf{97.3}        \\
SARSIM-M & 55.5 / \textbf{88.4}       & 50.8 / \textbf{90.5}        & 53.9 / \textbf{92.4}        & 60.3 / \textbf{93.5}        & 61.2 / \textbf{93.6}        \\
SARSIM-G & 44.8 / \textbf{87.9}        & 40.3 / \textbf{87.0}        & 44.2 / \textbf{88.1}        & 50.6 / \textbf{87.9}        & 52.5 / \textbf{87.2}        \\
FakeData     & 42.1 / \textbf{98.5}        & 27.0 / \textbf{99.5}        & 21.7 / \textbf{99.6}        & 21.6 / \textbf{99.5}        & 21.3 / \textbf{99.5}        \\
MNIST        & 15.1 / \textbf{94.1}        & 15.5 / \textbf{95.8}        & 16.6 / \textbf{96.3}        & 21.1 / \textbf{95.5}        & 22.2 / \textbf{95.8}        \\
\bottomrule
\end{tabular}
}
\end{table}

%% file: sections/S4_conclusion.tex
\section{Conclusion}

We now confirm our initial hypothesis (Sec.~\ref{sec:intro}) that modern SSL techniques like SimCLR are capable of learning highly generalizable SAR feature extractors from large pools of diverse and unlabeled SAR data. 
As evidence, our experiments show that a model trained mostly on unlabeled SAR ships is able to provide an informative-enough feature space to do few-shot classification of MSTAR targets.
This result provides merit to the under-studied global model approach to SAR ATR and confirms that SAR features can be highly transferable, even across sensors, imaging modes, target types, etc.
At a more concrete level, we provide a method for performing robust few-shot SAR ATR in a limited environment where there is no labeled data for pretraining.
We also provide a methodology that boosts OOD detection performance of the classifier without adding significant assumptions or overhead.
We hope that this work motivates further study on few-shot SAR ATR and representation learning for SAR, with specific considerations for robustness to OOD inputs and the trade-off between OOD detection and generalization.

Lastly, we provide some suggestions for future work.
We believe there are potential gains in developing improved domain-relevant augmentations for both Stage 1 and Stage 2 training. 
We also believe that adding synthetically generated data crafted to be relevant to the downstream task may improve the quality of learned features.
Next, one may investigate the use of different SSL algorithms and backbone model architectures. 
Finally, an interesting study would be to evaluate when it becomes beneficial to fine-tune the feature extractor instead of keeping it fixed during Stage 2 training.

\noindent \textbf{Public Release Number:}~AFRL-2022-3418

%% file: sections/appendix.tex
\clearpage
\onecolumn
\section*{Supplemental Materials}

%%%%%%%%%%%%%%%%%%%%%%%%%%%%%%%%%%%%%%%%%%%%%%%%%%%%% 
\subsection*{A. SimCLR Augmentations for SAR Data}

\begin{figure}[h]
\centering
\begin{minipage}{.52\textwidth}
    \vspace{7mm}
  \centering
  \includegraphics[width=\linewidth]{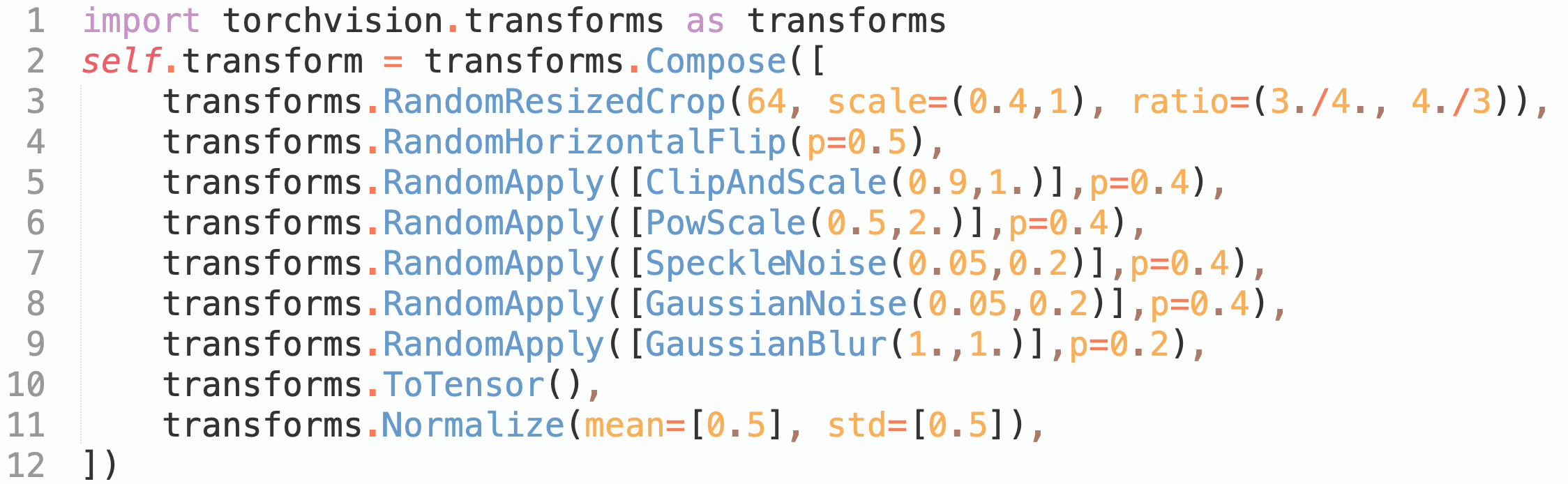}
  \vspace{1mm}
  \caption{PyTorch-like code describing transforms used to train SimCLR model.}
  \label{fig:app1}
\end{minipage}\hfill
\begin{minipage}{.45\textwidth}
  \centering
  \includegraphics[width=\linewidth]{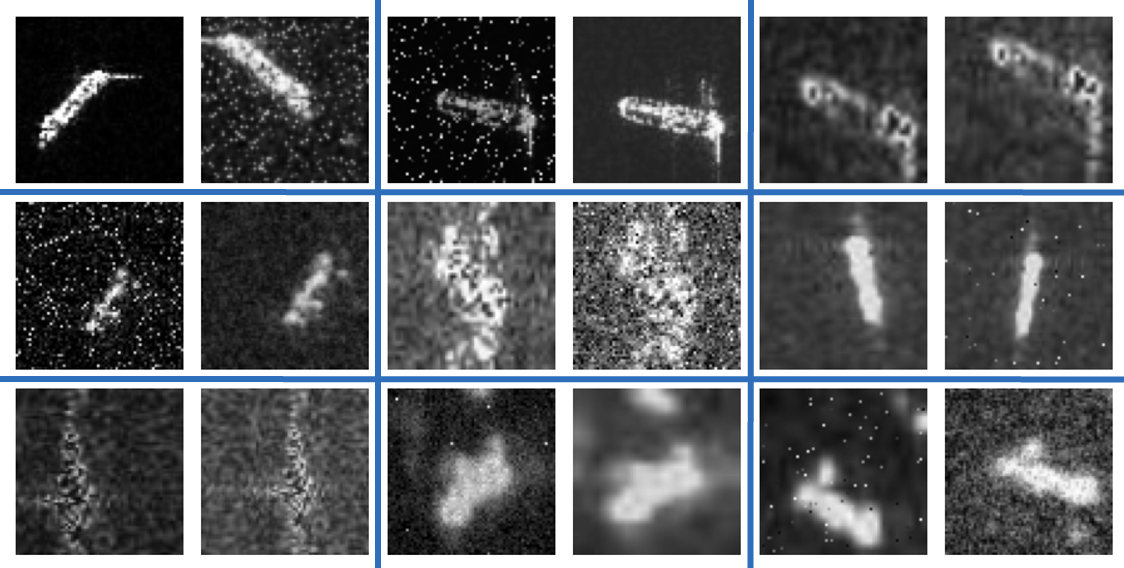}
  \caption{Examples of augmented pairs.}
  \label{fig:app2}
\end{minipage}
\end{figure}

In this section we will discuss the augmentation pipeline used to train the SimCLR model in more detail. 
Fig.~\ref{fig:app1} shows the PyTorch transform block that defines the augmentations.
Notice, several of the operations are from \texttt{torchvision.transforms} and we leave description of those to the official docs (\url{https://pytorch.org/vision/0.8/transforms.html}).
The remaining transforms are defined as follows:
\begin{itemize}
    \item \texttt{ClipAndScale(a,b)} - linearly re-scale the range of pixel values in the image given a randomly chosen maximum value between \texttt{a} and \texttt{b}. Pseudo: $\text{max\_val}=random.uniform(a,b); x_{new}=clip(x_{orig}, 0, \text{max\_val})/\text{max\_val}$. 
    \item \texttt{PowScale(a,b)} - pixel-wise exponentiation using a randomly selected value between \texttt{a} and \texttt{b}. Pseudo: $\text{val}=random.choice([random.uniform(a,1),random.uniform(1,b)]); x_{new}=(x_{orig})^\text{val}$. 
    \item \texttt{SpeckleNoise(a,b)} - randomly replace a subset of pixel values with values chosen from a Uniform(0,1) distribution. The size of the subset in \% total pixels is decided with $random.uniform(a,b)$.
    \item \texttt{GaussianNoise(a,b)} - apply pixel-wise additive Gaussian noise using $stdev=random.uniform(a,b)$.  
    \item \texttt{GaussianBlur} - convolve the chip with a Gaussian filter. 
\end{itemize}
Finally, Fig.~\ref{fig:app2} shows several examples of augmented pairs that are generated with this transform block. Recall, the SimCLR training objective encourages the two different views in the pair to lie near each-other in projection space, while views from different pairs should lie far away.

%%%%%%%%%%%%%%%%%%%%%%%%%%%%%%%%%%%%%%%%%%%%%%%%%%%%% 
\subsection*{B. Comparison With A Powerful Recent Method for Few-Shot MSTAR}

In the main document we do not compare our method to meta-learning-based FSL algorithms. 
As mentioned several times, this is because meta-learning approaches assume the existence of a labeled dataset for pretraining the feature extractor, which is a violation of our basic assumption set. 
However, for the sake of completeness we now present a detailed comparison with a state-of-the-art meta-learning approach to few-shot MSTAR called Domain Knowledge Powered Two Stream Deep Network (DKTS-N) \cite{dktsn}. Below are some notable points in no particular order:
\begin{itemize}
    \item DKTS-N uses the labeled SARSIM dataset \cite{sarsim_dataset} as the source of pretraining data for meta-learning. We argue that this is essentially an ideal scenario because SARSIM is specifically/conveniently synthesized to match the conditions of MSTAR (i.e., the downstream FSL task). We believe that this implies a more significant assumption than any we make in our work. Also, it is unclear if the method would be nearly as effective if the pretraining dataset was not so closely related to the downstream task. Finally, although our method is not reliant on such a related dataset, in practice we could leverage more data by adding it to $\mathcal{D}_{pretrain}$.
    \item DKTS-N reports results using a preprocessing function that performs perfect azimuth angle normalization on both the training and test data. We believe this is done using meta-data provided in the MSTAR dataset. They also show that as the normalization function incurs errors, performance can degrade significantly. It is unclear how one would achieve perfect azimuth angle normalization in the ``wild,'' and thus we do not use it here. However, if such a function were developed, we believe it could complement our model in a similar way to theirs.
    \item DKTS-N does not consider OOD detection in the few-shot classifier and thus is susceptible to producing erroneous predictions on OOD data at test time.
    \item DKTS-N involves a time-intensive inference procedure which scales poorly as \#-shots increases  (refer Table VII of \cite{dktsn}). This is because for each test input it performs an iterative optimization procedure w.r.t~each support sample. Our inference procedure is not nearly as complex, and is a single forward pass through a DNN (e.g., a RN18).
    \item While DKTS-N shows very high performance in MSTAR SOC, our method actually outperforms it in several of the EOC settings. Specifically, in the (4-way, 10-shot) and (4-way, 25-shot) cases our method is over 3\% more accurate. 
\end{itemize}

%%%%%%%%%%%%%%%%%%%%%%%%%%%%%%%%%%%%%%%%%%%%%%%%%%%%% 
\newpage
\subsection*{C. Additional results using a different self-supervised pretraining algorithm}

In the main manuscript we use the SimCLR algorithm to pretrain the feature extractor in Stage 1 of the pipeline. 
SimCLR is often used as a standard baseline for comparison in Self-Supervised Learning (SSL) literature and our intention is to use it as a representative of the SSL-class of algorithms.
To be clear, our framework is not tied to the SimCLR algorithm and we believe that any modern SSL algorithm may be used for pretraining.
To show the potential impact of a different SSL algorithm we run experiments with the Bootstrap Your Own Latent$^\dagger$ (BYOL) method.
BYOL uses a non-contrastive distillation scheme to perform the representation learning. 
This is thought to offer some advantages over SimCLR because the learning signal is not reliant on many negative samples to contrast the positive pair with.
Because the field of SSL is advancing so quickly it is unclear which SSL paradigm will ultimately emerge as the ``best,'' but for now these two methodologies make for an interesting comparison.
As mentioned in the paper, an important future work is still to consider using different SSL algorithms for pretraining within our framework.

Table~\ref{tab:mstar_soc_BYOL} shows the extended few-shot classification results in MSTAR Standard Operation Conditions, where the Scratch and SimCLR numbers are copy-pasted from Table~\ref{tab:mstar_soc} in the main paper.
The BYOL training parameters are nearly identical to the SimCLR parameters, including the extractor architecture, optimization procedure, augmentation scheme, and training schedule. 
The only method specific parameter is the EMA momentum which we set to $\tau=0.9995$. 
In the table we see that BYOL actually outperforms the Scratch and SimCLR models in all scenarios. 
Interestingly, the benefits of BYOL over SimCLR are most clear in the very low-shot scenarios (i.e., 1- and 5-shots) where the average margin of improvement is $\sim$5\%.
Focusing on the 10-way scenario, BYOL outperforms the best scratch model by at least 12\% in the 1-10 shot range.
Finally, we note that BYOL-based models trained with 10-shots perform on par with Scratch models trained with 20-shots!
The ease at which BYOL is swapped into our framework is a key advantage and speaks to the flexibility and extendability we were striving for.

\input{tables/table_mstar_soc_BYOL}

\vspace{4mm}
\noindent
[$\dagger$] Jean-Bastien Grill, Florian Strub, Florent Altché, Corentin Tallec, Pierre H. Richemond, Elena Buchatskaya, Carl Doersch, Bernardo Avila Pires, Zhaohan Daniel Guo, Mohammad Gheshlaghi Azar, Bilal Piot, Koray Kavukcuoglu, Rémi Munos, and Michal Valko. ``Bootstrap your own latent: A new approach to self-supervised learning.'' In \textit{Advances in Neural Information Processing Systems}, 2020.

% %%%%%%%%%%%%%%%%%%%%%%%%%%%%%%%%%%%%%%%%%%%%%%%%%%%%% 
% \subsection*{D. EXTRA RESULTS FOR MY EYES ONLY}

% \input{IEEEtran/tables/table_mstar_soc_EXTRA}

% \input{IEEEtran/tables/table_mstar_eoc_EXTRA}

%% file: tables/table_mstar_soc_BYOL.tex
%%%%%%%%%%%%%%%%%%%%%%%%%%%%%%%%%%%%%%%%%%%%%%% MSTAR SOC RESULTS -- NO ROTATED LABELS
\begin{table*}[h]
\centering
\caption{Extended MSTAR Standard Operating Condition Accuracy Results (with 95\% confidence interval)}
\vspace{-1mm}
\label{tab:mstar_soc_BYOL}
\resizebox{1.\textwidth}{!}{
\begin{tabular}{lccccccccccccccc}
\toprule
                 & \multicolumn{5}{c}{2-way}                     & \multicolumn{5}{c}{5-way}                     & \multicolumn{5}{c}{10-way}                    \\ \cmidrule(lr){2-6} \cmidrule(lr){7-11} \cmidrule(lr){12-16}
Method           & {1-shot} & {5-shot} & {10-shot} & {20-shot} & {25-shot} & {1-shot} & {5-shot} & {10-shot} & {20-shot} & {25-shot} & {1-shot} & {5-shot} & {10-shot} & {20-shot} & {25-shot} \\ \midrule
Scratch-AConv    &  59.6\scriptsize{$\pm$1.2}  & 73.6\scriptsize{$\pm$1.2}   & 84.3\scriptsize{$\pm$0.9}    & 92.1\scriptsize{$\pm$0.7}    & 93.8\scriptsize{$\pm$0.6}    & 31.6\scriptsize{$\pm$0.6}   & 55.9\scriptsize{$\pm$0.8}   & 72.3\scriptsize{$\pm$0.8}    & 85.7\scriptsize{$\pm$0.5}    & 88.5\scriptsize{$\pm$0.4}    & 20.9\scriptsize{$\pm$0.3}   & 47.4\scriptsize{$\pm$0.4}   & 65.0\scriptsize{$\pm$0.4}    & 80.7\scriptsize{$\pm$0.2}    & 84.5\scriptsize{$\pm$0.2}    \\
Scratch-RN18     &  55.8\scriptsize{$\pm$1.2}  & 74.9\scriptsize{$\pm$1.4}   & 84.1\scriptsize{$\pm$1.1}    & 89.6\scriptsize{$\pm$0.8}    & 92.1\scriptsize{$\pm$0.6}    & 30.6\scriptsize{$\pm$0.7}   & 53.7\scriptsize{$\pm$0.8}   & 67.0\scriptsize{$\pm$0.7}    & 79.3\scriptsize{$\pm$0.6}    & 83.2\scriptsize{$\pm$0.5}    & 20.7\scriptsize{$\pm$0.3}   & 43.4\scriptsize{$\pm$0.4}   & 58.9\scriptsize{$\pm$0.4}    & 74.0\scriptsize{$\pm$0.4}    & 78.5\scriptsize{$\pm$0.3}    \\ [0.2mm] \cdashline{1-16} \\[-2.3mm]
SimCLR+basic     &  64.5\scriptsize{$\pm$1.4}  & 83.0\scriptsize{$\pm$1.1}   & 91.4\scriptsize{$\pm$0.7}    & 96.1\scriptsize{$\pm$0.4}    & 97.4\scriptsize{$\pm$0.2}    & 38.8\scriptsize{$\pm$0.7}   & 68.4\scriptsize{$\pm$0.7}   & 82.6\scriptsize{$\pm$0.5}    & 91.8\scriptsize{$\pm$0.3}    & 94.0\scriptsize{$\pm$0.2}    & 28.3\scriptsize{$\pm$0.3}   & 59.2\scriptsize{$\pm$0.3}   & 75.2\scriptsize{$\pm$0.2}    & 87.7\scriptsize{$\pm$0.1}    & 90.6\scriptsize{$\pm$0.1}    \\
SimCLR+OE        &  65.4\scriptsize{$\pm$1.5}  & 86.4\scriptsize{$\pm$1.0}   & 92.2\scriptsize{$\pm$0.8}    & 96.5\scriptsize{$\pm$0.4}    & 97.5\scriptsize{$\pm$0.2}    & 39.9\scriptsize{$\pm$0.7}   & 69.7\scriptsize{$\pm$0.6}   & 83.3\scriptsize{$\pm$0.5}    & 92.3\scriptsize{$\pm$0.3}    & 93.8\scriptsize{$\pm$0.2}    & 28.8\scriptsize{$\pm$0.3}   & 60.5\scriptsize{$\pm$0.3}   & 75.6\scriptsize{$\pm$0.2}    & 87.0\scriptsize{$\pm$0.1}    & 89.3\scriptsize{$\pm$0.1}    \\
[0.2mm] \cdashline{1-16} \\[-2.3mm]
BYOL+basic        &  72.4\scriptsize{$\pm$1.9}  & 86.7\scriptsize{$\pm$1.1}   & 93.8\scriptsize{$\pm$0.6}    & 97.1\scriptsize{$\pm$0.3}    & 98.0\scriptsize{$\pm$0.2}    & 44.7\scriptsize{$\pm$0.9}   & 74.5\scriptsize{$\pm$0.6}   & 86.0\scriptsize{$\pm$0.5}    & 93.9\scriptsize{$\pm$0.2}    & 94.8\scriptsize{$\pm$0.2}    & 32.8\scriptsize{$\pm$0.4}   & 65.1\scriptsize{$\pm$0.3}   & 79.5\scriptsize{$\pm$0.2}    & 89.8\scriptsize{$\pm$0.1}    & 92.1\scriptsize{$\pm$0.1}    \\
BYOL+OE        &  70.1\scriptsize{$\pm$1.8}  & 87.6\scriptsize{$\pm$1.1}   & 94.7\scriptsize{$\pm$0.5}    & 97.8\scriptsize{$\pm$0.2}    & 98.1\scriptsize{$\pm$0.2}    & 45.9\scriptsize{$\pm$0.8}   & 75.9\scriptsize{$\pm$0.7}   & 87.4\scriptsize{$\pm$0.4}    & 93.8\scriptsize{$\pm$0.2}    & 94.9\scriptsize{$\pm$0.2}    & 33.7\scriptsize{$\pm$0.4}   & 66.2\scriptsize{$\pm$0.4}   & 79.8\scriptsize{$\pm$0.2}    & 88.7\scriptsize{$\pm$0.1}    & 90.7\scriptsize{$\pm$0.1}    \\
\bottomrule
\end{tabular}
}
\vspace{-2.5mm}
\end{table*}